%% file: arxiv.tex
\documentclass{article}

\usepackage[preprint, nonatbib]{neurips_2024}

\usepackage{wrapfig}

\usepackage[utf8]{inputenc} 
\usepackage[T1]{fontenc}    
\usepackage{hyperref}       
\usepackage{url}            
\usepackage{booktabs}       
\usepackage{amsfonts}       
\usepackage{nicefrac}       
\usepackage{microtype}      
\usepackage{xcolor}         
\usepackage{enumitem}
\usepackage{graphicx}
\usepackage{algorithm}
\usepackage{algpseudocode}
\usepackage{amsmath}
\usepackage[normalem]{ulem}
\useunder{\uline}{\ul}{}

\title{Unique3D: High-Quality and Efficient 3D Mesh Generation from a Single Image}

\newcommand{\eg}{\textit{e.g., }}
\newcommand{\ie}{\textit{i.e., }}

\author{%
    Kailu Wu \\
    Tsinghua University \\
    \And
    Fangfu Liu \\
    Tsinghua University \\
    \And
    Zhihan Cai \\
    Tsinghua University \\
    \And
    Runjie Yan \\
    Tsinghua University \\
    \And
    Hanyang Wang \\
    Tsinghua University \\
    \And
    Yating Hu \\
    AVAR Inc. \\
    \And
    Yueqi Duan\footnotemark[2] \\
    Tsinghua University \\
    \And
    Kaisheng Ma\footnotemark[2] \\
    Tsinghua University \\
}

\begin{document}

\maketitle

\renewcommand{\thefootnote}{\fnsymbol{footnote}}
\footnotetext[2]{Corresponding Author}

\begin{figure}[h]
\begin{center}
    \includegraphics[width=1.0\linewidth]{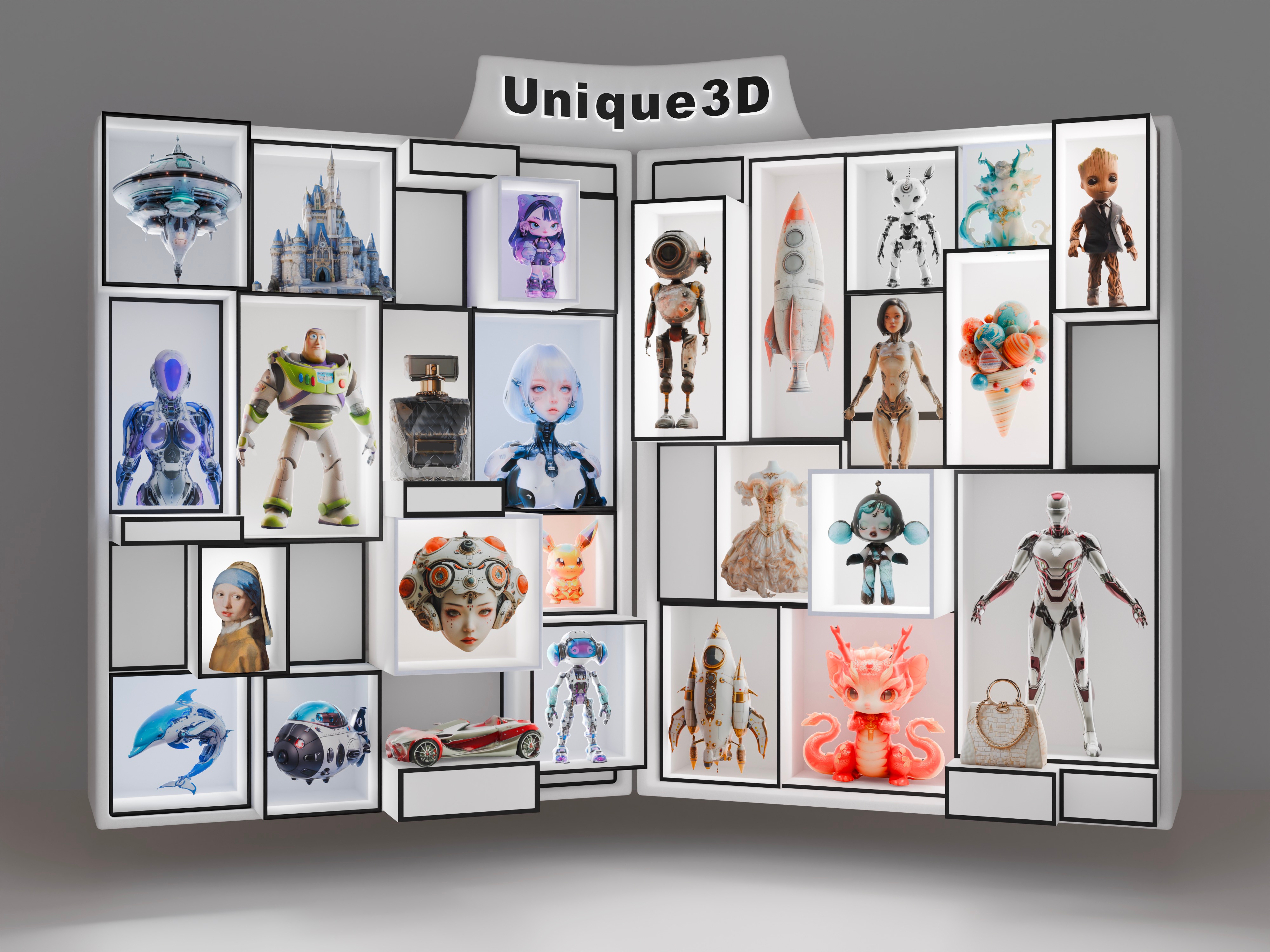}
\end{center}
\caption{\textbf{Gallery of Unique3D}. High-fidelity and diverse textured mesh generated by Unique3D from single-view wild images within 30 seconds. \href{https://wukailu.github.io/Unique3D/}{Project Link}.}
\label{fig:teaser}
\end{figure}

\input{nips2024/0-abstract}
\input{nips2024/1-introduction}
\input{nips2024/2-related-work}

\input{nips2024/3-method}

\input{nips2024/4-experiments}
\input{nips2024/5-conclusion}

\medskip
{
\small
\bibliographystyle{unsrt}
\bibliography{reference}
}

\input{nips2024/6-appendix}

\end{document}

%% file: nips2024/0-abstract.tex
\begin{abstract}
    In this work, we introduce \textbf{Unique3D}, a novel image-to-3D framework for efficiently generating high-quality 3D meshes from single-view images, featuring state-of-the-art generation fidelity and strong generalizability. Previous methods based on Score Distillation Sampling (SDS) can produce diversified 3D results by distilling 3D knowledge from large 2D diffusion model, but they usually suffer from long per-case optimization time with inconsistent issues. Recent works address the problem and generate better 3D results either by finetuning a multi-view diffusion model or training a fast feed-forward model. However, they still lack intricate textures and complex geometries due to inconsistency and limited generated resolution. To simultaneously achieve high fidelity, consistency, and efficiency in single image-to-3D, we propose a novel framework Unique3D that includes a multi-view diffusion model with a corresponding normal diffusion model to generate multi-view images with their normal maps, a multi-level upscale process to progressively improve the resolution of generated orthographic multi-views, as well as an \underline{i}n\underline{s}tant and c\underline{o}nsistent \underline{me}sh \underline{r}econstruction algorithm called \textit{ISOMER}, which fully integrates the color and geometric priors into mesh results. Extensive experiments demonstrate that our Unique3D significantly outperforms other image-to-3D baselines in terms of geometric and textural details. Project page: \url{https://wukailu.github.io/Unique3D/.} 
\end{abstract}

%% file: nips2024/1-introduction.tex
\section{Introduction}
Automatically generating diverse and high-quality 3D content from single-view images is a fundamental task in 3D Computer Vision~\cite{jun2023shap-E, LRM, long2023wonder3d, CRM, xu2024instantmesh}, which can facilitate a wide range of versatile applications~\cite{liu2024-3D-gen-survey, 3DSurvey}, including gaming, architecture, art, and animation. However, this task is challenging and ill-posed due to the underlying ambiguity of 3D geometry in a single view. 

Recently, the rapid development of diffusion models~\cite{DDPM, DDIM, stable-diffusion} has opened up new perspectives for 3D content creation. Powered by the strong prior of 2D image diffusion models, DreamFusion~\cite{dreamfusion} proposes Score Distillation Sampling (SDS) to address the limitation of 3D data by distilling 3D knowledge from 2D diffusions~\cite{Imagen}, inspiring the progress of SDS-based 2D lifting methods~\cite{magic3d, qian2023magic123, prolific-dreamer, liu2023sherpa3d, fantasia3D}. Despite their diversified compelling results, they usually suffer from long per-case optimization time for hours, poor geometry, and inconsistent issues (\eg, Janus problem~\cite{dreamfusion}), thus not practical for real-world applications. To overcome the problems, a series of works leverage larger-scale open-world 3D datasets~\cite{Objaverse, chang2015shapenet, deitke2024objaverse-xl} either to fine-tune a multi-view diffusion model~\cite{long2023wonder3d, liu2023syncdreamer, ImageDream} and recover the 3D shapes from the generated multi-view images or train a large reconstruction model (LRM)~\cite{LRM, GRM, CRM, xu2024instantmesh} by directly mapping image tokens into 3D representations (\eg, triplane or 3D Gaussian~\cite{3DGS}). However, due to local inconsistency in mesh optimization~\cite{long2023wonder3d, yang2024magic-boost} and limited resolution of the generative process with expensive computational overhead~\cite{LRM, xu2024instantmesh}, they struggle to produce intricate textures and complex geometric details with high resolution.

In this paper, we present a novel image-to-3D framework for efficient 3D mesh generation, coined \textbf{Unique3D}, to address the above challenges and simultaneously achieve high-fidelity, consistency, and generalizability. Given an input image, Unique3D first generates orthographic multi-view images from a multi-view diffusion model. Then we introduce a multi-level upscale strategy to progressively improve the resolution of generated multi-view images with their corresponding normal maps from a normal diffusion model. Finally, we propose an \underline{i}n\underline{s}tant and c\underline{o}nsistent \underline{me}sh \underline{r}econstruction (\textit{ISOMER}) algorithm to reconstruct high-quality 3D meshes from the multiple RGB images and normal maps, which fully integrates the color and geometric priors into mesh results. 
Both diffusion models are trained on a filtered version of the Objaverse dataset~\cite{Objaverse} with $\sim 50k$ 3D data. To enhance the quality and robustness, we design a series of strategies into our framework, including the noise offset channel in the multi-view diffusion training process to correct the discrepancy between training and inference~\cite{lin2024common-zero-snr}, a stricter dataset filtering policy, and an expansion regularization to avoid normal collapse in mesh reconstruction. Overall, our method can generate high-fidelity, diverse, and multi-view consistent meshes from single-view wild images within 30 seconds, as shown in Figure~\ref{fig:teaser}.

We conduct extensive experiments on various wild 2D images with different styles. The experiments verify the efficacy of our framework and show that our Unique3D outperforms existing methods for high fidelity, geometric details, high resolution, and strong generalizability. 

In summary, our contributions are:
\begin{itemize}[leftmargin=*]

\item We propose a novel image-to-3D framework called Unique3D that holistically archives a leading level of high-fidelity, efficiency, and generalizability among current methods.

\item We introduce a multi-level upscale strategy to progressively generate higher-resolution RGB images with the corresponding normal maps.

\item We design a novel instant and consistent mesh reconstruction algorithm (\textit{ISOMER}) to reconstruct 3D meshes with intricate geometric details and texture from RGB images and normal maps.

\item Extensive experiments on image-to-3D tasks demonstrate the efficacy and generation fidelity of our method, unlocking new possibilities for real-world deployment in the field of 3D generative AI.

\end{itemize}

%% file: nips2024/2-related-work.tex
\section{Related Work}
\textbf{Mesh Reconstruction}.
Despite the significant advancements in various 3D representations (\eg, SDF~\cite{SDF, DeepSDF}, NeRF~\cite{NeRF, InstantNGP}, 3D Gaussian~\cite{3DGS}), meshes remain the most widely used 3D format in popular 3D engines (\eg, Blender, Maya) with a mature rendering pipeline. Reconstructing high-quality 3D meshes efficiently from multi-view or single-view images is a daunting task in graphics and 3D computer vision. Early approaches usually adopt a laborious and complex photogrammetry pipeline with multiple stages, with techniques like Structure from motion (SfM)~\cite{agarwal2011building-rome-in-a-day, schonberger2016structure-from-motion-revisited, snavely2006photo-tourism}, Multi-View Stereo (MVS)~\cite{furukawa2015multi-view-stereo, schonberger2016pixelwise}, and mesh surface extraction~\cite{poisson, lorensen1998marching}. Powered by deep learning and powerful GPUs, recent works~\cite{DMTet, FlexiCubes, huang2018deepmvs, yao2018mvsnet, LRM, CRM, GRM} have been proposed to pursue higher efficiency and quality with gradient-based mesh optimization or even training a large feed-forward reconstruction network. However, their pipeline still suffers from heavy computational costs and struggles to adapt to complex geometry. To balance efficiency and quality, we propose a novel instant and high-quality mesh reconstruction algorithm in this paper that can reconstruct complex 3D meshes with intricate geometric details from sparse views.

\textbf{Score Distillation for 3D Generation}.
Recently, data-driven large-scale 2D diffusion models have achieved notable success in image and video generation~\cite{stable-diffusion, Imagen, yang2023diffusion, singer2022make-a-video}. However, transferring it to 3D generation is non-trivial due to curating large-scale 3D datasets. Pioneering works DreamFusion~\cite{dreamfusion} proposes Score Distillation Sampling (SDS) (also known as Score Jacobian Chaining~\cite{SJC}) to distill 3D geometry and appearance from pretrained 2D diffusion models when rendered from different viewpoints. The following works continue to enhance various aspects such as fidelity, prompt alignment, consistency, and further applications~\cite{magic3d, qian2023magic123, prolific-dreamer, fantasia3D, li2023sweetdreamer, tang2023dreamgaussian, ye2024dreamreward}. However, such optimization-based 2D lifting methods are limited by long per-case optimization time and multi-face problem~\cite{MVDream} due to lack of explicit 3D prior. As Zero123~\cite{Zero123} proves that Stable Diffusion~\cite{stable-diffusion} can be finetuned to generate novel views by conditioning on relative camera poses, one-2-3-45~\cite{liu2024one-2-3-45} directly produce plausible 3D shapes from generated images in Zero123. Though it achieves high efficiency, the generated results show poor quality with a lack of texture details and 3D consistency.

\textbf{Multi-view Diffusion Models for 3D Generation}.
To achieve efficient and 3D consistent results, some works~\cite{long2023wonder3d, liu2023syncdreamer, Zero123Plus, ImageDream, MVDream} fine-tune the 2D diffusion models with large-scale 3D data~\cite{Objaverse} to generate multi-view consistent images and then create 3D contents using sparse view reconstruction. For example, SyncDreamer~\cite{liu2023syncdreamer} leverages attention layers to produce consistent multi-view color images and then use NeuS~\cite{wang2021neus} for reconstruction. Wonder3D~\cite{long2023wonder3d} explicitly encodes the geometric information into 3D results and improves quality by cross-domain diffusion. Although these methods generate reasonable results, they are still limited by local inconsistency from multi-views generated by out-domain input images and limited generated resolution from the architecture design, producing coarse results without high-resolution textures and geometries. In contrast, our method can generate higher-quality textured 3D meshes with more complex geometric details within just 30 seconds.

%% file: nips2024/3-method.tex
\section{Method}
\begin{figure}[!t]
    \small
    \centering
    \includegraphics[width=\linewidth]{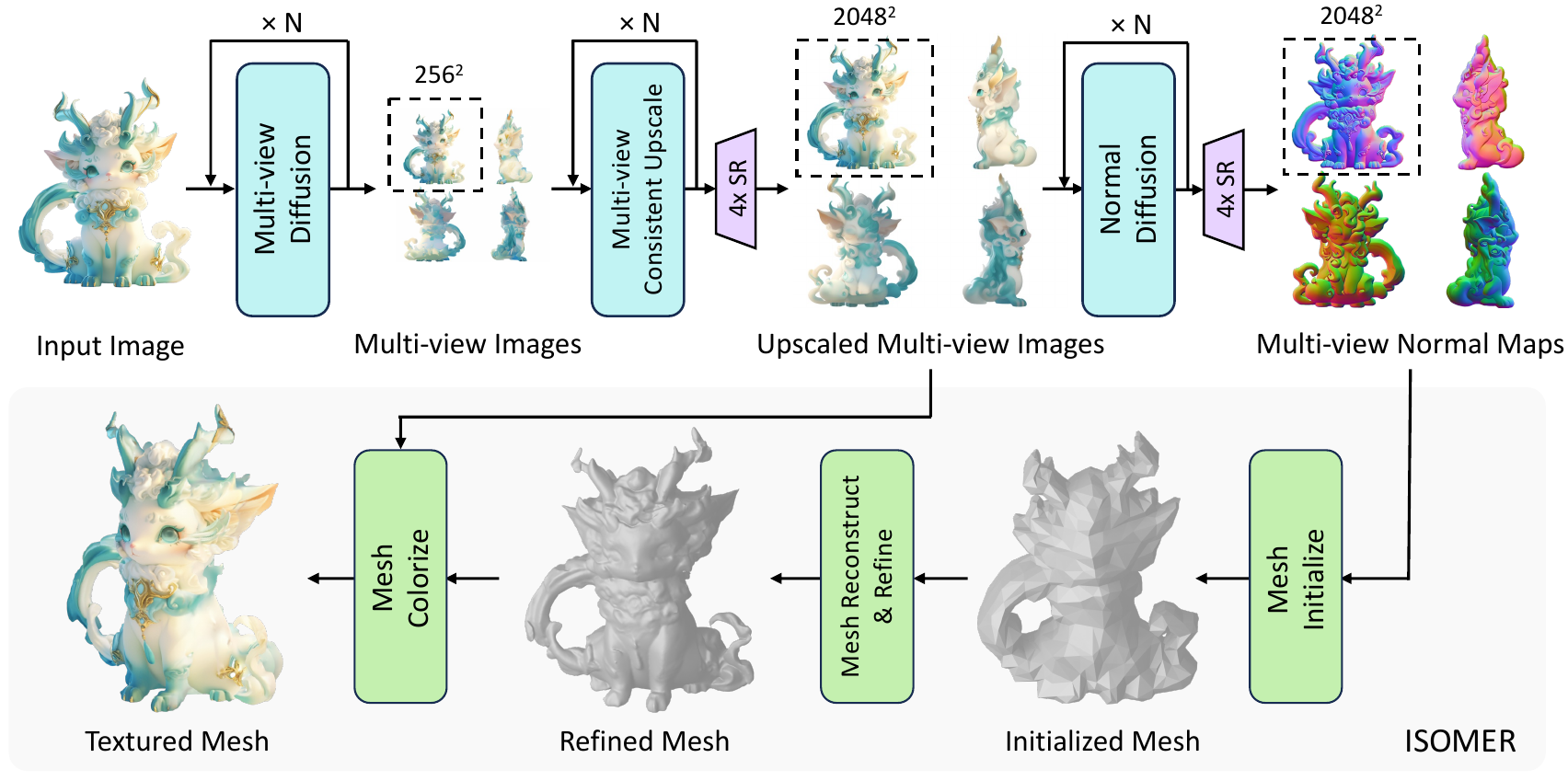}
    \vspace{-0.25cm}
    \caption{\textbf{Pipeline of our Unique3D}. Given a single in-the-wild image as input, we first generate four orthographic multi-view images from a multi-view diffusion model. Then, we progressively improve the resolution of generated multi-views through a multi-level upscale process. Given generated color images, we train a normal diffusion model to generate normal maps corresponding to multi-view images and utilize a similar strategy to lift it to high-resolution space. Finally, we reconstruct high-quality 3D meshes from high-resolution color images and normal maps with our instant and consistent mesh reconstruction algorithm \textit{ISOMER}.}
    \vspace{-0.5cm}
    \label{fig:method_overview}
\end{figure}

In this section, we introduce our framework, \ie Unique3D, for high-fidelity, efficient, and generalizable 3D mesh generation from a single in-the-wild image. Given an input image, we first generate four orthographic multi-view images with their corresponding normal maps from a multi-view diffusion model and a normal diffusion model. Then, we lift them to high-resolution space progressively, (Sec~\ref{subsec: high-resolution-generation}). Given high-resolution multi-view RGB images and normal maps, we finally reconstruct high-quality 3D meshes with our instant and consistent mesh reconstruction algorithm \textit{ISOMER}, (Sec~\ref{subsec: ISOMER}). ISOMER directly handles the case where the global normal of the same vertex is inconsistent across viewpoints to enhance the consistency.
An overview of our framework is depicted in Figure~\ref{fig:method_overview}.

\subsection{High-resolution Multi-view Generation}
\label{subsec: high-resolution-generation}
We first explain the design of our high-resolution multi-view generation model that generates four orthographic view images from a single input image. Instead of directly training a high-resolution (2K) multi-view diffusion that would consume excessive computational resources, we adopt a multi-level generation strategy to upscale the generated resolution progressively.

\textbf{High-resolution Multi-view Image Generation}.
Instead of training from scratch, we start with the initialization of the pre-trained 2D diffusion model using the checkpoint of Stable Diffusion~\cite{LDM} and encode multi-view dependencies to fine-tune it to obtain a multi-view diffusion model that is able to generate four orthographic view images (256 resolution) from a single in-the-wild image. It is worth noting that the images generated in this step have relatively low resolution and suffer from multi-view inconsistency in out-of-the-domain data. This significantly limits the quality of recent works~\cite{GRM, CRM, xu2024instantmesh, long2023wonder3d, Zero123Plus}. In contrast, we address the multi-view consistency issue during the reconstruction phase (Sec~\ref{subsec: ISOMER}). Given the generated four orthographic view images, we then finetune a multi-view aware ControlNet~\cite{ControlNet} to improve the resolution of images. This model leverages the four collocated RGB images as control information to generate corresponding clearer and more precise multi-view results. It enhances the details and ameliorates unclear regions, leading the resolution of images from 256 to 512. Finally, we employ a single-view super-resolution model~\cite{RealESRGAN} to further upscale the image by a factor of four, achieving a resolution of 2048 that offers sharper edges and details without disrupting the multi-view consistency.

\textbf{High-resolution Normal Map Prediction}. Using pure RGB images makes it extremely hard to reconstruct correct geometry. To effectively capture the rich surface details of the target 3D shape, we finetune a normal diffusion model to predict normal maps corresponding to multi-view color images. Similar to the above high-resolution image generation stage, we also employ the super-resolution model~\cite{RealESRGAN} to quadruple the normal resolution, which enables our method to recover high-fidelity geometric details, especially the accuracy of the edges.

To enhance the capability of the image generation model and the standard normal prediction model in producing high-quality images with uniform backgrounds, we adopt a channel-wise noise offset strategy~\cite{ZeroSNR}. This can alleviate the problem caused by the discrepancy between the initial Gaussian noise during sampling and the noisiest training sample.

\subsection{ISOMER: An Efficient Method for Direct Mesh Reconstruction}
\label{subsec: ISOMER}
Despite impressive results generated by recent popular image-to-3D methods~\cite{long2023wonder3d, liu2024make-your-3d, xu2024instantmesh, LRM, CRM} that follow the field-based reconstruction~\cite{DMTet, FlexiCubes, DiffMC}, they have limited potential for higher-resolution applications as their computational load is proportional to the cube of the spatial resolution. \textit{In contrast, we design a novel reconstruction algorithm directly based on mesh, where the computational load scales with only the square of the spatial resolution and relates to the number of faces, thus achieving a fundamental improvement.} This enables our model to efficiently reconstruct meshes with tens of millions of faces within seconds.

We now move to introduce our instant and consistent mesh reconstruction algorithm (\textit{ISOMER}), which is a robust, accurate, and efficient approach for direct mesh reconstruction from high-resolution multi-view images. Specifically, the \textit{ISOMER} consists of three main steps: (a) estimating the rough topological structure of the 3D object and generating an initial mesh directly; (b) employing a coarse-to-fine strategy to further approximate the target shape; (c) explicitly addressing inconsistency across multiple views to reconstruct high-fidelity and intricate details. 
Notably, the entire mesh reconstruction process takes no more than 10 seconds.

\textbf{Initial Mesh Estimation}. 
Unlike popular reconstruction methods based on signed distance fields~\cite{InstantNSR} or occupancy fields~\cite{NeRF}, mesh-based reconstruction methods~\cite{DS, NDS} struggle with changing topological connectivity during optimization, which requires correct topological construction during initialization. Although initial mesh estimation can be obtained by existing methods like DMTet~\cite{DMTet}, they cannot accurately reconstruct precise details (\eg, small holes or gaps). To address the problem, we utilize front and back views to directly estimate the initial mesh, which is fast for accurate recovery of all topologically connected components visible from the front. Specifically, we 
integrate the normal map from the frontal view to obtain a depth map by 
\vspace{-0.2cm}
\begin{equation}
    d(i, j)=\sum_t=0^i n_x(t, j)
\end{equation}
where $n_x(t, j)$ is the normal vector of the $t$-th pixel in the $j$-th row.
Although the diffusion process generates pseudo normal maps, these maps do not yield a real normal field which is irrotational. To address this, we introduce a random rotation to the normal map before integration. The process is repeated several times, and the mean value of these integrations is then utilized to calculate the depth, providing a reliable estimation.
Subsequently, we map each pixel to its respective spatial location using the estimated depth, creating mesh models from both the front and back views of the object. The two models are seamlessly joined through Poisson reconstruction, which guarantees a smooth connection between them. Finally, we simplify them into $2000$ fewer faces for our mesh initialization.

\textbf{Coarse-to-Fine Mesh Optimization}.
Building upon the research in inverse rendering \cite{Remesh, ContinuesRemeshing, NVdiffrast}, we iteratively optimize the mesh model to minimize a loss function. During each optimization step, the mesh undergoes differentiable rendering to compute the loss and gradients, followed by vertex movement according to the gradients. Finally, the mesh is corrected after iteration through edge collapse, edge split, and edge flip to maintain a uniform face distribution and reasonable edge lengths. After several hundred coarse-to-fine iterations, the model converges to a rough approximation of the target object's shape. The loss function for this part includes a mask-based loss 
\begin{equation}
    \mathcal{L}_{mask} = \sum_i \left \| \hat{M_i} - M^{pred}_i \right \|_2^2,  
\end{equation}
where $\hat{M_i}$ is the rendered mask under view $i$ and $M^{pred}_i$ is the predicted mask from previous subsection under view $i$. The mask-based loss regulates the mesh contour. Additionally, it includes a normal-based loss 
\begin{equation}
    \mathcal{L}_{normal} = \sum_i M^{pred}_i \otimes \left \| \hat{N_i} - N^{pred}_i \right \|_2^2,
\end{equation}
concerning the rendered normal map $\hat{N_i}$ of the object and the predicted normal map $N^{pred}_i$, optimizing the normal direction in the visible areas, where $\otimes$ denotes element-wise production. We compute the final loss function as:  
\begin{equation}
    \mathcal{L}_{recon} = \mathcal{L}_{mask} + \mathcal{L}_{normal}.   
\end{equation}

To address potential surface collapse issues under limited-view normal supervision as shown in Figure~\ref{fig:ablation}-(b), we employ a regularization method called Expansion. At each step, vertices are moved a small distance in the direction of their normals, akin to weight decay. 

\textbf{ExplicitTarget Optimization for Multi-view Inconsistency and Geometric Refinement}. Due to inherent inconsistencies in generated multi-view images from out-of-distribution (OOD) in-the-wild input, no solution can perfectly align with every viewpoint. After the above steps, we can only reconstruct a model that roughly matches the shape but lacks detail, falling short of our pursuit of high-quality mesh. Therefore, we cannot use the common method that minimizes differences in all views, which would lead to significant wave-pattern flaws, as shown in Figure~\ref{fig:ablation}-(a). 
To overcome this challenge, finding a more suitable optimization target becomes crucial. Under single-view supervision, although a complete model cannot be reconstructed, the mesh shape within the visible area of that view can meet the supervision requirements with highly detailed structures. Based on this, we propose a novel method that assigns a unique optimization target for each vertex to guide the optimization direction. In contrast to the conventional implicit use of multi-view images as optimization targets, we \textbf{explicitly} define the optimization target with better robustness. We call this explicit optimization target as \textit{ExplicitTarget} and devise it as follows:

(\textit{ExplicitTarget}). Let  $Avg(V, W) = \frac{\sum_i{V_i W_i}}{W_i}$ represent the weighted average function, and $V_{M}(v, i): (\mathbb{N}^+, \mathbb{N}^+) \rightarrow \{0, 1\}$ represent the visibility of vertex $v$ in mesh $M$ under view $i$. $Col_M(v,i) $ Indicate the color of vertex v in viewpoint i.  We compute the ExplicitTarget $ET$ of each vertex in mesh $M$ as
\begin{equation}
    ET_{M}(v) = \begin{cases}
Avg(Col_M(v,i), V_{M}(v, i) W_M(v, i)^2) 
 & \text{, if } \sum_{i} V_{M}(v, \mathcal{i}) > 0 \\ 
 \mathbf{0} & \text{, otherwise, } 
\end{cases} 
\end{equation}
where$W_M(v, i) = -\cos(N_{v}^{(M)}, N_{i}^{(view)})$ is a weighting factor that$N_{v}^{(M)}$ is the vertex normal of $v$ in mesh $M$, and $N_{i}^{(view)}$ is the view direction of view $i$. 



In the function $ET_{M}(\mathcal{I}, \mathcal{I}_m)$, the predicted color of vertex $v$ is computed as the weighted sum of supervised views, with weights determined by the square of cosine angles. This is because the projected area is directly proportional to the cosine value, and the prediction accuracy is also positively correlated with the cosine value. The object loss function for ExplicitTarget is defined as 
\vspace{-0.2cm}
\begin{equation}
    \mathcal{L}_{ET} = \sum_i M^{pred}_i \otimes \left \| \hat{N_i} - N^{ET}_{i}\right \|_2^2,
\end{equation}
where $N^{ET}_{i}$ is the rendering result of mesh $M$ with $\{ET_{M}(\mathcal{I}, N^{pred}, v)| v\in M\}$ under the $i$-th viewpoint. The final optimization loss function is 
\begin{equation}
    \mathcal{L}_{refine} = \mathcal{L}_{mask} + \mathcal{L}_{ET}.
\end{equation}
Towards this end, we finish the introduction of the \textit{ISOMER} reconstruction process, which includes three stages: Initialization, Reconstruction, and Refinement.

Upon generating precise geometric structures, it is necessary to colorize them based on multi-view images. Given the inconsistencies across multi-view images, the colorizing process adopts the same method used in the refinement stage. Specifically, the colors of 
mesh $M$ is $\{ET_{M}(\mathcal{I}, \mathcal{I}_{rgb}^{pred}, v)| v\in M\}$. Moreover, certain regions of the model may remain unobservable from the multi-view perspective, necessitating the coloring of these invisible areas. To address this, we utilize an efficient smoothing coloring algorithm to complete the task. More detailed and specific algorithmic procedures can be found in the Appendix.

%% file: nips2024/4-experiments.tex
\section{Experiments}

\begin{figure}[t!]
\begin{center}
    \includegraphics[width=0.98\linewidth]{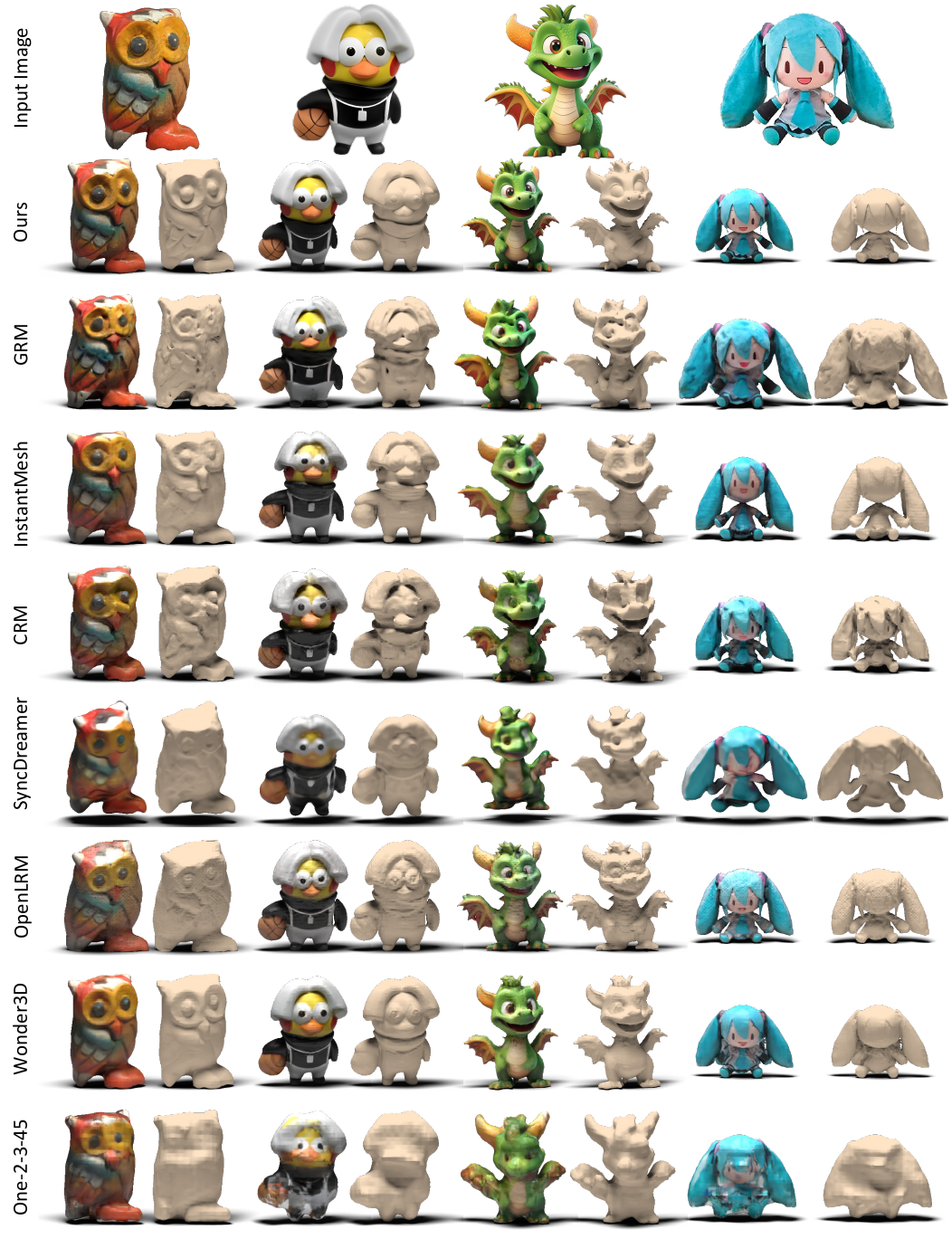}
\end{center}
\vspace{-0.5cm}
\caption{\textbf{Qualitative Comparison}. Our approach provides superior geometry and texture.}
\label{fig:exp_main}
\vspace{-1.0cm}
\end{figure}

\subsection{Experimental Setting}

\textbf{Dataset:} Utilizing a subset of the Objaverse dataset as delineated by LGM~\cite{LGM}, we apply a rigorous filtration process to exclude scenes containing multiple objects, low-resolution imagery, and unidirectional faces, leading to a refined dataset of approximately $50k$ objects. 
To address surfaces without thickness, we render eight orthographic projections around each object horizontally. By examining the epipolar lines corresponding to each horizontal ray, we identify $13k$ instances of illegitimate data.
For rendering, we employ random environment maps and lighting to augment the dataset, thereby enhancing the model’s robustness. To ensure high-quality generation, all images are rendered at a resolution of $2048 \times 2048$ pixels.

\textbf{Network Architecture:} The initial level of image generation is initialized with the weight of the Stable Diffusion Image Variations Model~\cite{LDM}, while the subsequent level employs an upscaled version fine-tuned from ControlNet-Tile~\cite{ControlNet}. The final stage uses the pre-trained Real-ESRGAN model~\cite{RealESRGAN}. Similarly, the initial stage of normal map prediction is initialized from the aforementioned Stable Diffusion Image Variations. Details of these networks are provided in the Appendix.

\textbf{Reconstruction Details:} The preliminary mesh structure is inferred from a normal map with a resolution of $256 \times 256$, which is then simplified to a mesh comprising $2,000$ faces. The reconstruction process involves $300$ iterations using the SGD optimizer~\cite{SGD}, with a learning rate of $0.3$. The weight of expansion regularization is set to $0.1$. Subsequent refinement takes $100$ iterations, maintaining the same optimization parameters.

\textbf{Training Details:} The entire training takes around 4 days on 8 NVIDIA RTX4090 GPUs. The primary level of multiview image generation uses $30k$ training iterations with a batch size of $1,024$. The training of multi-view image upscaling involves $10k$ iterations with a batch size of $128$. Normal map prediction is trained for $10k$ iterations at a batch size of $128$. Additional training specifics are accessible in the Appendix.

\subsection{Comparisons}

\begin{figure}[t]
\begin{center}
    \includegraphics[width=0.95\linewidth]{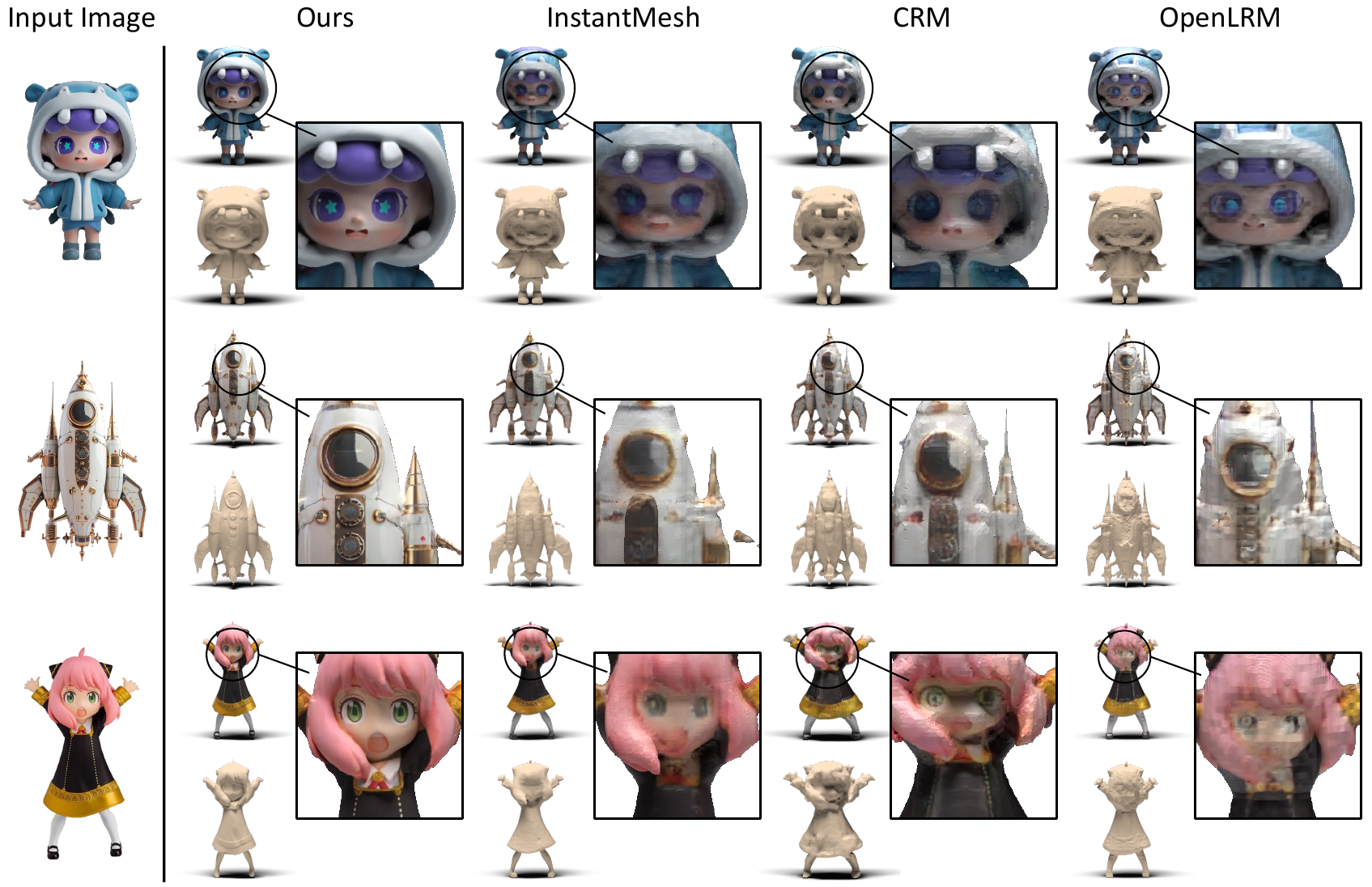}
\end{center}
\caption{\textbf{Detailed Comparison}. We compare our model with InstantMesh~\cite{xu2024instantmesh}, CRM~\cite{CRM} and OpenLRM~\cite{openlrm}. Our models generates accurate geometry and detailed texture.}
\label{fig:exp_detail}
\end{figure}

\textbf{Qualitative Comparison:}
To highlight the advantages of our methodology, we perform a comprehensive comparison with existing works, including CRM~\cite{CRM}, one-2-3-45~\cite{liu2024one-2-3-45}, Wonder3D~\cite{long2023wonder3d}, OpenLRM~\cite{openlrm}, and InstantMesh~\cite{xu2024instantmesh}. 
For a fair quality comparison, we choose to present samples previously selected in the referenced papers, originating from Wonder3D~\cite{long2023wonder3d}, SyncDreamer~\cite{liu2023syncdreamer}, CRM~\cite{CRM}, and InstantMesh~\cite{xu2024instantmesh}. The results are shown in Figure~\ref{fig:exp_main}. Our results clearly surpass the existing works in both geometric and material quality, thereby emphasizing the benefits of our approach in achieving high resolution and intricate details in both geometry and material. In addition to the above overall quality comparison, we further show the comparison of the details in Figure~\ref{fig:exp_detail}, highlighting the advantage of our method in high resolution. The reconstruction process of ISOMER is completed in under $10$ seconds, while the entire procedure from the input image to high-precision mesh is accomplished in less than $30$ seconds on an RTX4090.

\begin{table}
    \caption{Quantitative comparison results for mesh visual and geometry quality. We report the metrics of PSNR, SSIM, LPIPS and Clip-Similarity~\cite{CLIP}, ChamferDistance (CD), Volume IoU and F-score on GSO~\cite{GSO} dataset.}
  \label{tab:quant_full}
  \centering
  \small
  \begin{tabular}{lccccccc}
  \toprule
  \textbf{Method} & \textbf{PSNR↑} & \textbf{SSIM↑}  & \textbf{LPIPS↓} & \textbf{Clip-Sim↑} & \textbf{CD↓} & \textbf{Vol. IoU↑} & \textbf{F-Score↑} \\
  \midrule 
  One-2-3-45      & 16.1058          & 0.8874          & 0.1812          & 0.7782             & 0.0313                  & 0.4142             & 0.5518            \\
  OpenLRM         & 18.0433          & 0.8957          & 0.1560          & 0.8416             & 0.0336                  & 0.3947             & 0.5354            \\
  Wonder3D        & 18.0932          & 0.8995          & 0.1536          & 0.8535             & 0.0261                  & 0.4663             & 0.6016            \\
  InstantMesh     & {\ul 18.8262}    & {\ul 0.9111}    & {\ul 0.1283}    & \textbf{0.8795}    & 0.0161                  & 0.5083             & 0.6491            \\
  CRM             & 18.4407          & 0.9088          & 0.1366          & 0.8639             & \textbf{0.0141}         & {\ul 0.5218}       & {\ul 0.6574}      \\
  \midrule 
  Unique3D        & \textbf{20.0611} & \textbf{0.9222} & \textbf{0.1070} & {\ul 0.8787}       & {\ul 0.0143}            & \textbf{0.5416}    & \textbf{0.6696}   \\
  \midrule 
  Unique3D w/o ET & 20.0383          & 0.9199          & 0.1129          & 0.8675             & 0.0158                  & 0.5320             & 0.6594            \\
  \midrule 
  Wonder3D+ISOMER     & 18.6131          & 0.9026          & 0.1470          & 0.8621             & 0.0244                  & 0.4743             & 0.6088           \\
  \bottomrule
  \end{tabular}
\end{table}

\textbf{Quantitative Comparison:}
In line with previous work, we evaluate our results using the Google Scanned Objects (GSO)~\cite{GSO} dataset. We render frontal views at a resolution of $1024\times 1024$ with Blender EEVEE as input for all methods. All generated mesh results are normalized to the bounding box $[-0.5, 0.5]$ to ensure alignment. The geometric quality is assessed by calculating the distance to the ground truth mesh using metrics such as Chamfer Distance (CD), Volume IoU, and F-Score. Concurrently, we render $24$ views around the object, selecting one of $[0, 15, 30]$ for elevation angles and $8$ evenly distributed azimuth angles spanning a full 360-degree rotation. We employ PSNR, SSIM, LPIPS, and Clip-Similarity~\cite{CLIP} to evaluate the visual quality. The results are presented in Table~\ref{tab:quant_full}. As evidenced in table, both our geometric and material quality outperform those of existing methods. We find that ISOMER can even be used to improve the consistency of other methods. For example, in Table~\ref{tab:quant_full}, we replace Wonder3D's reconstruction method with ISOMER, which is not only faster but also of higher quality.

\subsection{Ablation Study and Disscussion}

\begin{figure}[t]
\small
\begin{center}
    \includegraphics[width=1.0\linewidth]{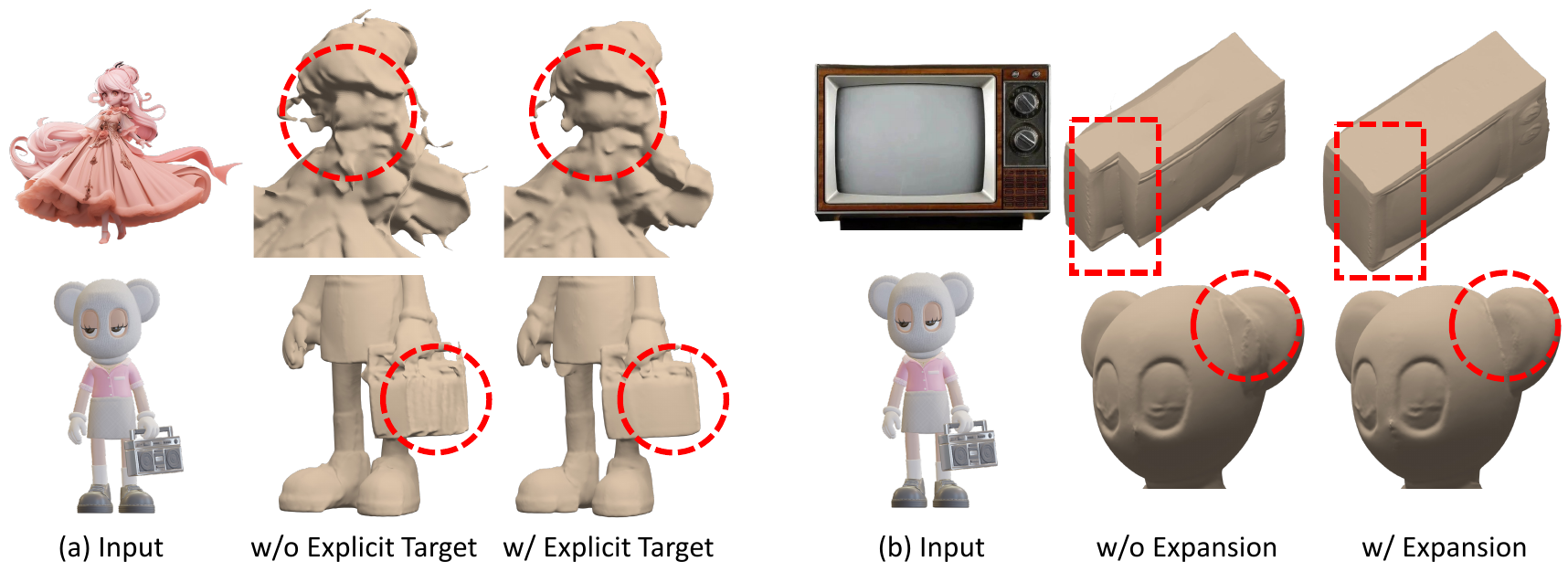}
\end{center}
\vspace{-0.25cm}
\caption{\textbf{Ablation Study on ISOMER}. (a) Without ExplicitTarget, the output mesh result has obvious defects. (b) Without expansion regularization, the output result collapses in some cases.}
\label{fig:ablation}
\vspace{-0.5cm}
\end{figure}

We analyze the importance of ExplicitTarget and expansion regularization in ISOMER. We compare samples with and without ExplicitTarget and Expansion Regularization in figure~\ref{fig:ablation}. We clearly show the improvement of ExplicitTarget for geometry and the necessity of expansion regularization for reconstruction. ExplicitTarget notably improves reconstruction results in challenging cases, while expansion regularization avoids some possible collapses.

\begin{figure}[t!]
\begin{center}
    \includegraphics[width=0.9\linewidth]{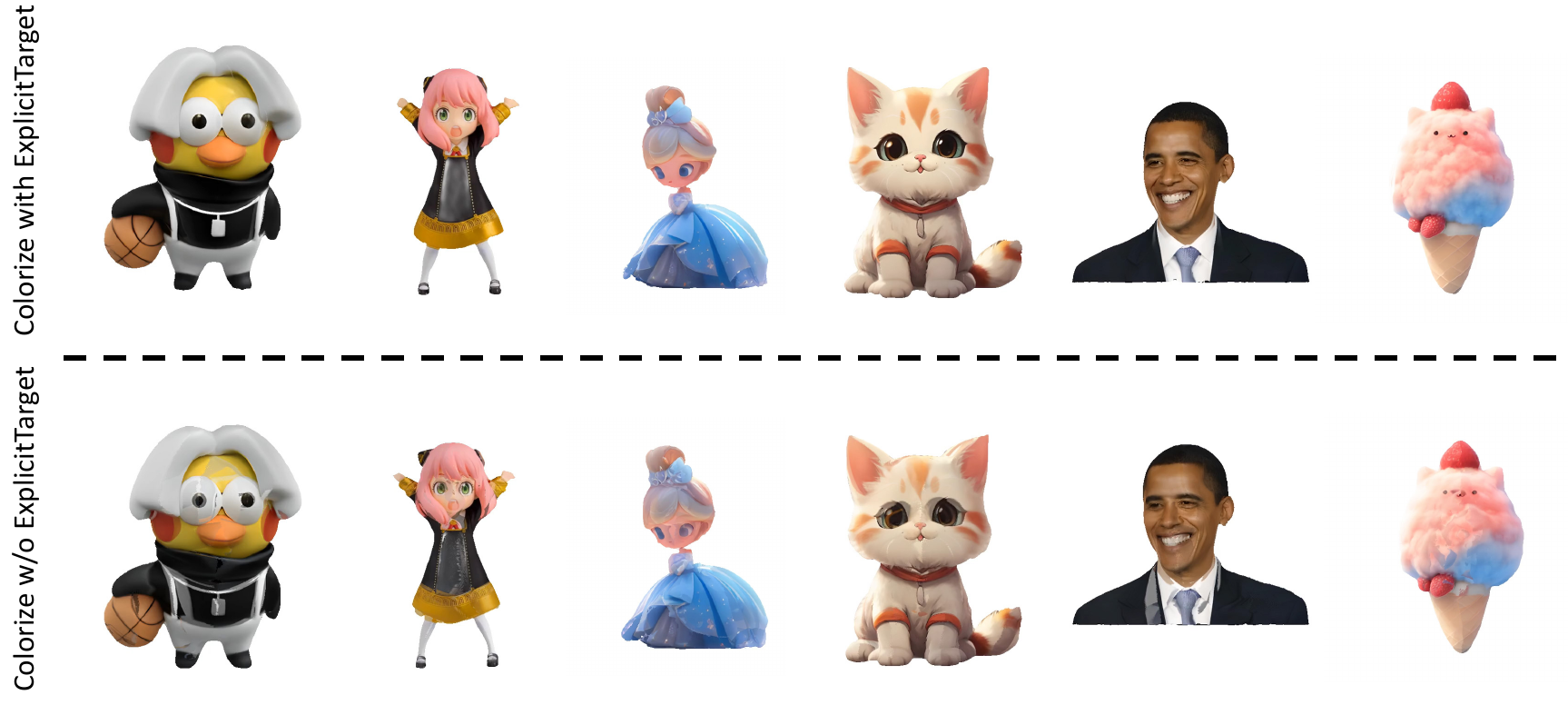}
\end{center}
\vspace{-0.5cm}
\caption{\textbf{Ablation on Colorize}. We show a comparison of whether or not to apply ExplicitTarget in coloring, and we can see that the group that does not use ExplicitTarget has significant artifacts, as there is no precise consistency across multiple views.}
\label{fig:exp_colorize}
\vspace{-0.25cm}
\end{figure}
Mirroring the approach used for geometry, we will include additional experimental results in Figure~\ref{fig:exp_colorize} that illustrate the impact of the Explicit Target method on texture quality. Without the Explicit Target, the results are obviously flawed.

\begin{table}
    \caption{Quantitative comparison results for ablation on 100 random samples with random rotation on GSO dataset. }
\label{tab:quant_rot}
\centering
\small
\begin{tabular}{lccccccc}
\toprule
\textbf{Method} & \textbf{PSNR↑} & \textbf{SSIM↑}  & \textbf{LPIPS↓} & \textbf{Clip-Sim↑} & \textbf{CD↓} & \textbf{Vol. IoU↑} & \textbf{F-Score↑} \\
\midrule
Unique3D        & 19.6744        & 0.9217         & 0.1101          & 0.8864             & 0.0118                  & 0.5463             & 0.6833            \\
\bottomrule
\end{tabular}
\end{table}
Additionally, we added a new test with randomly rotated objects sampled from $azimuth \in U[-180, 180], elevation \in U[-30, 30]$ in Table~\ref{tab:quant_rot} to test robustness in non-front-facing views. The test results show that Unique3D still performs well in this case, and even the geometry prediction is more accurate.

\begin{figure}[t!]
\begin{center}
    \includegraphics[width=0.8\linewidth]{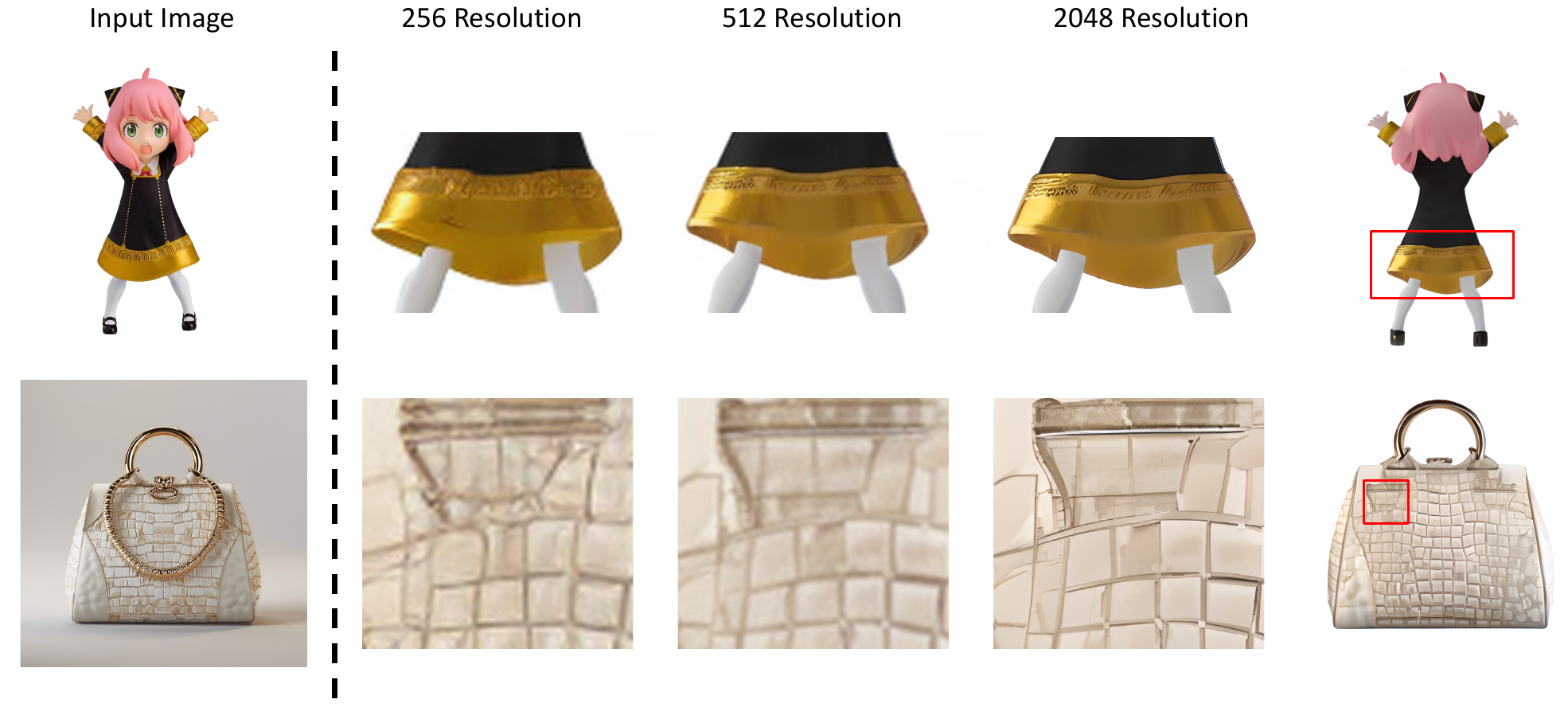}
\end{center}
\vspace{-0.5cm}
\caption{\textbf{Ablation on Resolution}. The visualization of the generated multi-views images at different stages is shown. Multi-level super-resolution does not change the general structure, but only improves the detail resolution, allowing the model to remain well-detailed.}
\label{fig:exp_resolution}
\vspace{-0.25cm}
\end{figure}
We expand our study to include a qualitative comparison across various resolutions in order to demonstrate the differences between different resolutions in Figure~\ref{fig:exp_resolution}. The results demonstrate the necessity of high resolution maps in generating high resolution meshes.

\begin{wrapfigure}{r}{0.54\linewidth}
    \vspace{-0.5em}
    \includegraphics[width=1.0\linewidth]{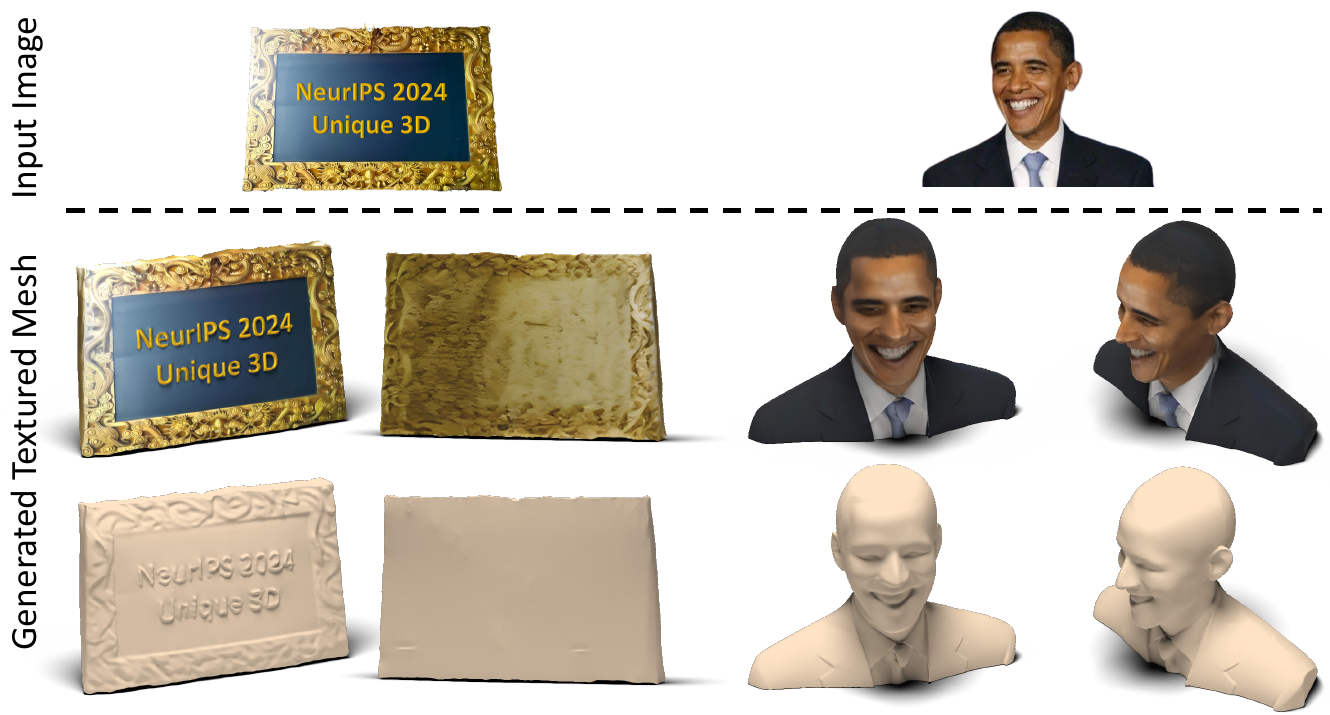}
    \vspace{-2em}
    \caption{\textbf{Challenging examples.}}
    \label{fig:challenge}
    \vspace{-1em}
\end{wrapfigure}

\textbf{Challenging Examples:}
The majority of existing common samples are overly simplistic to effectively demonstrate the advantages of our study. Consequently, we select two complex samples: an object featuring detailed text and a photograph of a human, as shown in Figure~\ref{fig:challenge}. It is observable that our method exhibits exceptional mesh materials and geometry, even capable of sculpting geometric structures with textual detail. In the context of photographs, our reconstruction results are nearly on par with specialized image-to-character mesh generation methods.

%% file: nips2024/5-conclusion.tex
\section{Conclusion}

In this paper, we introduce Unique3D, a pioneering image-to-3D framework that efficiently generates high-quality 3D meshes from single-view images with unprecedented fidelity and consistency. By integrating advanced diffusion models and the powerful reconstruction method ISOMER, Unique3D generates detailed and textured meshes within $30$ seconds, significantly advancing the state-of-the-art in 3D content creation from single images.

\textbf{Limitation and Future Works.} Our method, while capable of generating high-fidelity textured meshes rapidly, faces challenges. The multi-view prediction model may produce less satisfactory predictions for skewed or non-perspective inputs. Furthermore, the geometric coloring algorithm currently does not support texture maps. In the future, we aim to enhance the robustness of the multi-view prediction model by training on a more extensive and diverse dataset.

%% file: nips2024/6-appendix.tex
\newpage

\appendix

\begin{figure}[h]
\begin{center}
    \includegraphics[width=0.97\linewidth]{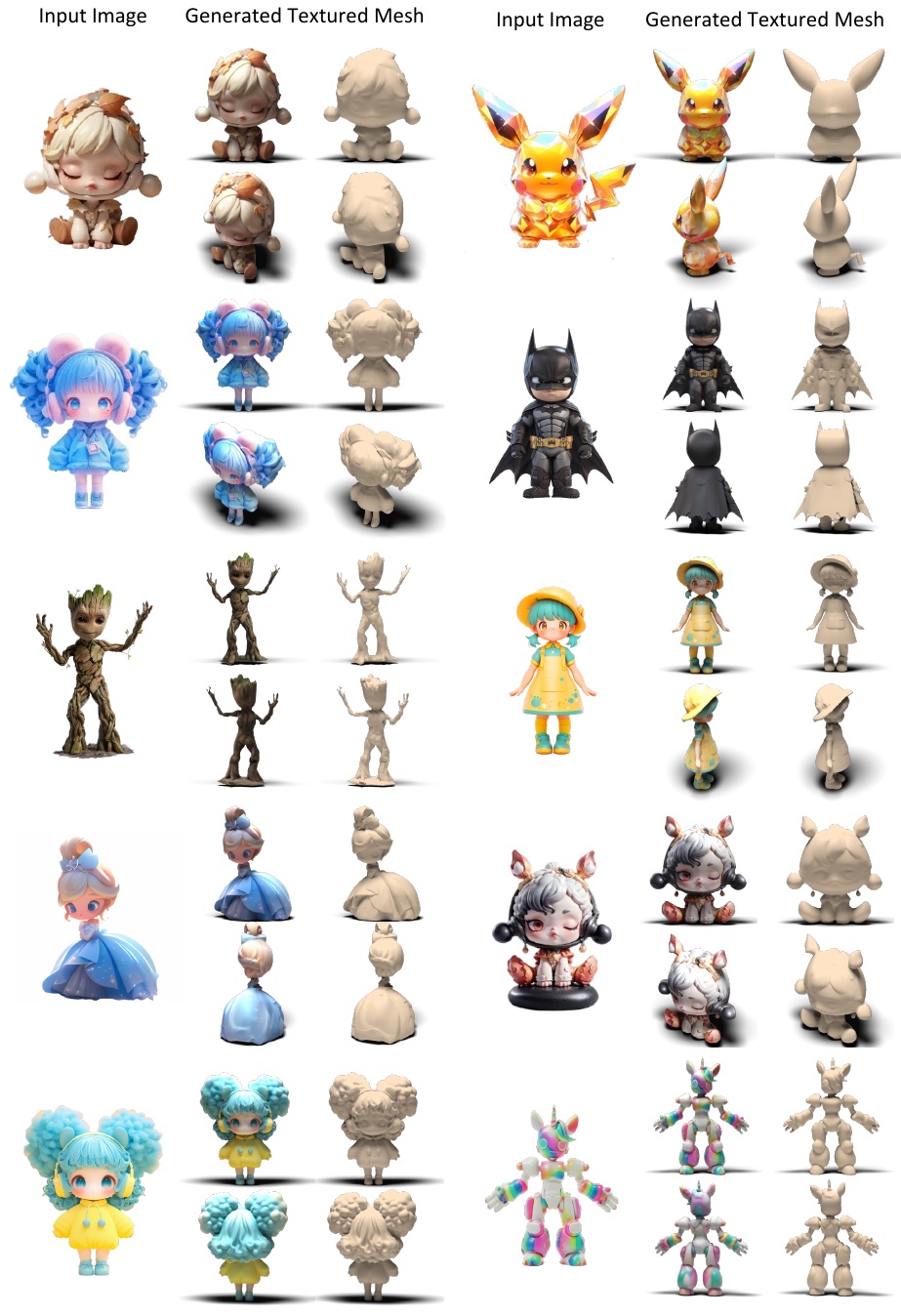}
\end{center}
\caption{More generated results of our method from a single image.}
\label{fig:more}
\vspace{-2em}
\end{figure}

\section{More Results}

We provide more generation results of our method from a single image in Figure~\ref{fig:more}.

\section{Network Architecture and Training Details}

\textbf{Multi-view Image Generation:}
In this part, we develop a model based on the architecture of Stable Diffusion Image Variation~\cite{LDM} with two main modifications: 
(1). The use of a class embedding, which takes an integer from 0 to 3 as input, indicating the corresponding view indexes.
(2). The simultaneous forward of four perspectives, where they are concatenated in the self-attention layers to achieve multi-view consistency.

For the training of the network, we utilize the following parameters:
\begin{itemize}
    \item A learning rate of $10^{-4}$.
    \item A batch size of $1024$.
    \item A noise offset of $0.1$.
    \item An SNR gamma of $5.0$.
    \item An 8-bit Adam optimizer~\cite{adam} with betas set to $(0.9, 0.999)$.
    \item An Adam weight decay of $0.01.$
    \item An Adam epsilon of $10^{-8}$.
    \item Gradient clipping with a norm of $1$ to ensure training stability.
\end{itemize}

\textbf{Multi-view Image Upscale:}
In this part, we aim to upscale merged low-resolution four-view images to a high resolution of 1024 pixels. To achieve this, we fine-tune the ControlNet-Tile network, leveraging StableDiffusion 1.5 as its backbone. Unlike traditional methods, we do not use text input; instead, we feed an empty text. Concurrently, we pass the input image through an IP-Adapter~\cite{ipadapter}. This approach allows the network to be guided in enhancing the multi-view details and achieving the desired resolution.

For the training of this network, we use the following parameters:
\begin{itemize}
    \item A learning rate of $5 \times 10^{-6}$.
    \item A batch size of $128$.
    \item A noise offset of $0.1$.
    \item An SNR gamma of $5.0$.
    \item An 8-bit Adam optimizer~\cite{adam} with betas set to $(0.9, 0.999)$.
    \item An Adam weight decay of $0.01.$
    \item An Adam epsilon of $10^{-8}$.
    \item Gradient clipping with a norm of $1$ to ensure training stability.
    \item Freeze parameters except for the ControlNet.
\end{itemize}

\textbf{Normal Prediction Diffusion:}
In this part, we train a diffusion model that takes an RGB image as input and produces its corresponding normal map as output. This model is based on the Stable Diffusion Image Variation~\cite{LDM}, with one key modification: a reference U-Net has been incorporated, which has an identical network structure and initialization as the original network. This reference U-Net provides pixel-wise reference attention to the main network exclusively at a new attention layer that added to the self-attention.

For the training of the network, we utilize the following parameters:
\begin{itemize}
    \item A learning rate of $10^{-4}$ for main network.
    \item A learning rate of $10^{-5}$ for reference network.
    \item A batch size of $128$.
    \item A noise offset of $0.1$.
    \item An SNR gamma of $5.0$.
    \item An 8-bit Adam optimizer~\cite{adam} with betas set to $(0.9, 0.999)$.
    \item An Adam weight decay of $0.01$.
    \item An Adam epsilon of $10^{-8}$.
    \item Gradient clipping with a norm of $1$ to ensure training stability.
    \item Freeze parameters except for the self-attention in reference attention.
    \item Train all parameters in the main network.
\end{itemize}

\section{Efficient Invisible Region Color Completion Algorithm}

In our approach to mesh coloring using multiple viewpoints, we encounter a minor yet noteworthy challenge: the need to color regions that are not directly visible. Although these regions are typically sparse and inconspicuous. In field-based representations such as Signed Distance Fields~\cite{SDF}, they often are the color of neighboring visible areas upon completion of the field optimization. To address this, we opt for a straightforward yet efficient algorithm that seamlessly spreads the colors of nearby visible regions into the invisible ones.

Our methodology employs a straightforward, multi-step color propagation algorithm, which stands out for its simplicity, swift execution, and reliability in delivering a reasonably detailed and nuanced color complement. This approach outperforms more complex, resource-intensive, and less stable techniques like using pre-trained inpainting diffusion models. The algorithm leverages the surrounding colors to gently fill in the invisible regions, with the detailed process outlined in Algorithm~\ref{alg:color}.

A critical aspect of the algorithm to consider is the potential for a stark color demarcation line if the process is halted immediately after all nodes have been colored. For example, in a one-dimensional scenario, if red is on the left and blue is on the right, separated by an uncolored section, stopping immediately will result in a high-contrast boundary. To mitigate this, we extend the color propagation process through a number of iterations to ensure a smooth color gradient. This allows the colors to gradually permeate throughout the entire connected component of the mesh that requires coloring, thus achieving a harmonious and visually coherent result.

\begin{algorithm}
    \small
    \caption{Color Completion Algorithm}
    \label{alg:color}
    \hspace*{\algorithmicindent} \textbf{Input:} Mesh $M$, list of invisible vertices $Inv$, list of color of all vertices $C$ \\
    \hspace*{\algorithmicindent} \textbf{Output:} The completed color list $C$
    \begin{algorithmic}[1]
    \State $cnt \gets 0$
    \State $stage2 \gets False$
    \State $colored \gets \emptyset$
    \ForAll{vertices $v$ in $M$}
        \If{$v \notin Inv$}
            \State Append $True$ to $visible\_vertices$
        \Else
            \State Append $False$ to $visible\_vertices$
        \EndIf
    \EndFor
    \While{$stage2 == False$ or $cnt > 0$}
        \ForAll{$i$ in $Inv$}
            \State $colored\_neighbors \gets$ list of vertices directly connected to $i$ in $M$ that have $colored == True$
            \If{$colored\_neighbors != \emptyset$ }
                \State $colored[i] \gets True$
                \State $C[i] \gets \text{mean}(C[colored\_neighbors])$
            \Else
                \State $colored[i] \gets False$
            \EndIf
        \EndFor
        \If{all elements of $colored$ are $True$}
            \State $stage2 \gets True$
            \State $cnt \gets cnt - 1$
        \Else
            \State $cnt \gets cnt + 1$
        \EndIf
    \EndWhile
    \State \Return $C$
    \end{algorithmic}
\end{algorithm}

\section{ExplicitTarget algorithm}

\begin{figure}[h]
\begin{center}
    \includegraphics[width=0.5\linewidth]{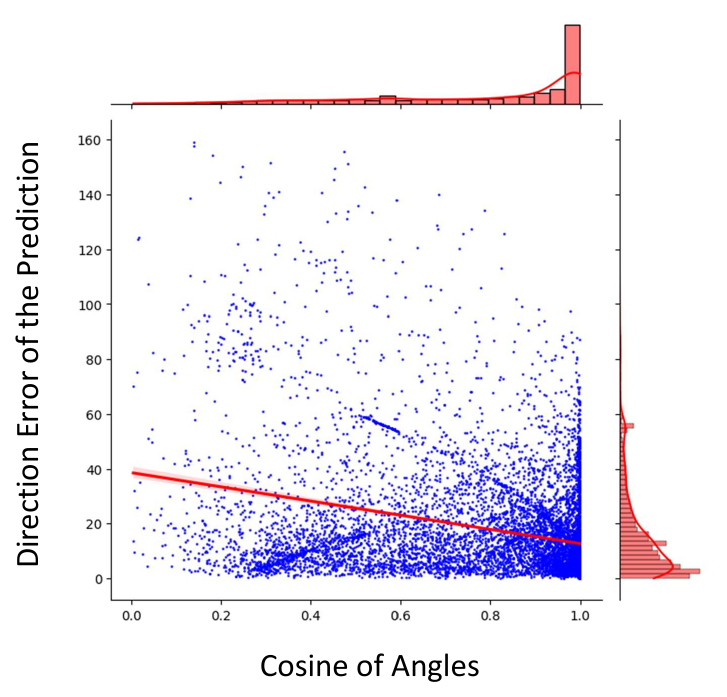}
\end{center}
\caption{Correlation between prediction value and prediction errors.}
\label{fig:cosine}
\end{figure}

In Algorithm~\ref{algo:explicit}, we demonstrate the detailed computation of ExplicitTarget. Specifically, we set an optimization target for each vertex, which is a weighted sum of the supervised signals from the visible views of the vertex. The weights are determined by two factors: the projected area of the nearby surface and the confidence level in the accuracy of the normals, which are used to calculate the weights. In Figure~\ref{fig:cosine}, we show the relationship between the normal results predicted by multiView diffusion and the accuracy of the predictions on the Objaverse~\cite{Objaverse} validation set. The results indicate that the closer the angle between the predicted normal and the vertical to the current viewpoint is, the lower the accuracy of the prediction. There is a negative correlation between these two factors, with a Pearson correlation coefficient of -0.304.

\begin{algorithm}[t]
    \caption{ExplicitTarget Algorithm}
    \label{algo:explicit}
    \hspace*{\algorithmicindent} \textbf{Input:} Multi-view image list $imgs$, initial mesh model $M$ \\
    \hspace*{\algorithmicindent} \textbf{Output:} Model $M'$ with vertex colors set to \textbf{ExplicitTarget}
    \begin{algorithmic}[1]
    \State $M' \gets M$
    \State Set the color of $M'$ to the vertex normals of $M'$
    \ForAll{vertices $v$ in the vertex set of $M'$}
        \State $tot\_weight \gets 0$
        \State $tot\_color \gets \vec{0}$ \Comment{Initialize to zero vector}
        \ForAll{images $im$ in $imgs$}
            \If{vertex $v$ is not visible in the viewpoint of $im$}
                \State \textbf{continue}
            \EndIf
            \State $ci \gets$ the color of vertex $v$ in image $im$
            \State $wi \gets$ the square of the cosine of the angle between the vertex normal of $v$ and the view direction from $im$ to $v$
            \State $tot\_weight \gets tot\_weight + wi$
            \State $tot\_color \gets tot\_color + wi \cdot \vec{ci}$
        \EndFor
        \If{$tot\_weight > 0$}
            \State Set the color of vertex $v$ in $M'$ to $tot\_color / tot\_weight$
        \EndIf
    \EndFor
    \State \Return $M'$
    \end{algorithmic}
\end{algorithm}

\section{More on Mesh Optimizations}
\textit{Edge Collapse}: This operation is used to avoid and heal defects in the mesh. It involves selecting an edge within a triangle and collapsing it to the other edge, effectively merging the two triangles into a single triangle. This process can help to eliminate narrow triangles that might be causing issues in the mesh, such as those that are too thin to accurately represent the surface they are approximating. Edge collapse can prevent the creation of topological artifacts and maintain the quality of the mesh.
\textit{Edge Split}: This is the opposite of edge collapse. In edge split, an edge that is longer than a specified maximum length is divided into two, creating new vertices at the midpoint of the edge. This operation is used to refine the mesh, ensuring that the local edge length is kept close to the optimal length. It helps to maintain the quality of the mesh by avoiding edges that are too long, which could lead to an inaccurate representation of the surface.
\textit{Edge Flip}: Edge flip is an operation that adjusts the connectivity of the mesh to improve its quality. It involves flipping an edge within a triangle to connect two non-adjacent vertices, effectively changing the triangulation of the mesh. This can help to maintain the degree of the vertices close to their optimal value, which is typically six for internal vertices (or four for boundary vertices).
The goal of these operations is to improve the mesh quality while avoiding defects and ensuring that the mesh accurately represents the target geometry. 

\section{Ablation Study on Mesh Initialization}

\begin{figure}[h]
\label{fig:albation_init}
\begin{center}
    \includegraphics[width=1.0\linewidth]{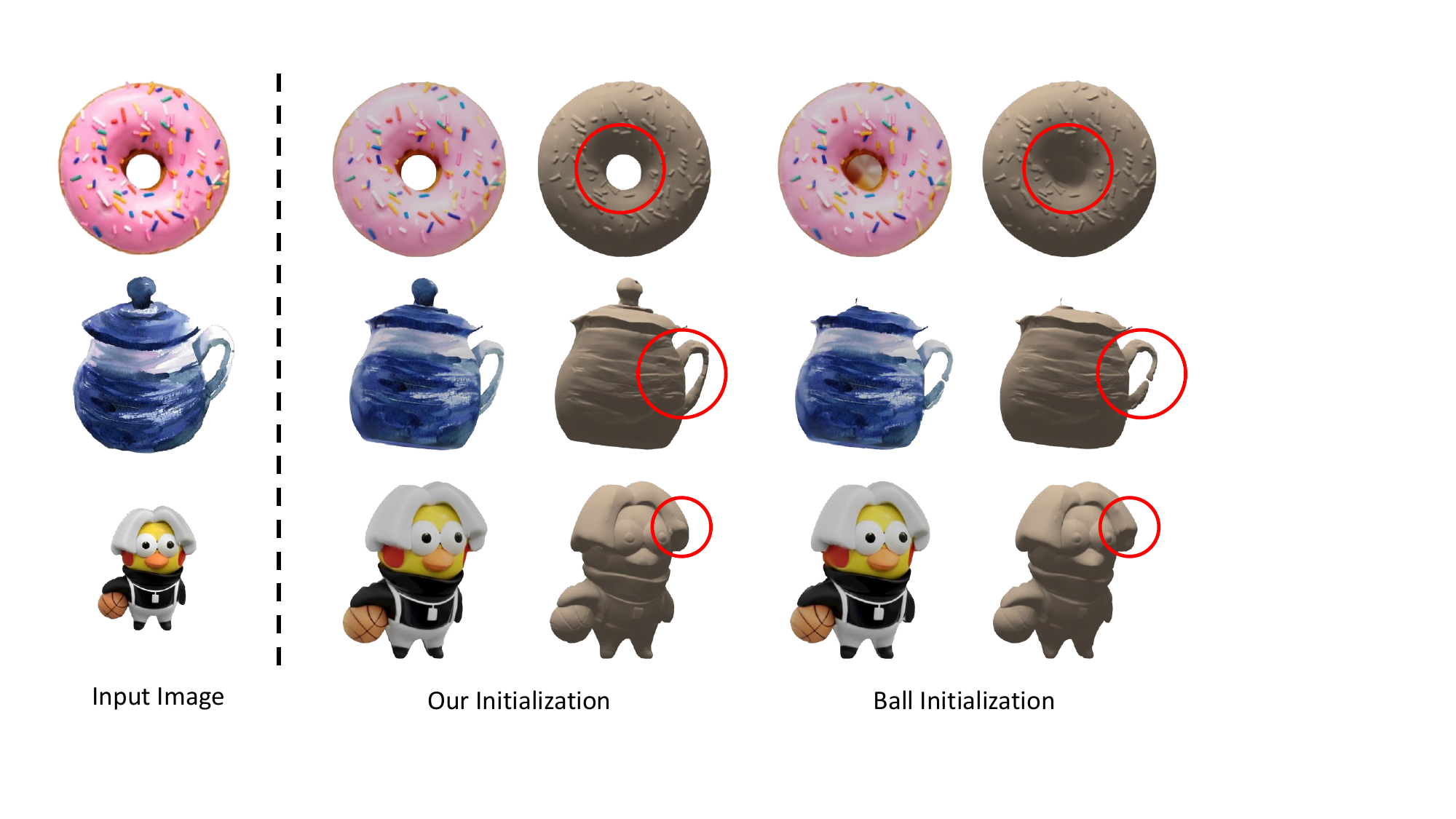}
\end{center}
\caption{\textbf{Ablations on Mesh Initialization}. We compare the results of using our fast initialization method, versus using a sphere as an initialization.}
\end{figure}

We compare the different mesh initialization methods and their results. One is our proposed fast initialization method, and the other is using spheres as initialization objects, a common practice in mesh-based reconstruction techniques. Figure~\ref{fig:albation_init} illustrates the problem of the mesh reconstruction method that fails to modify its topological structure. For example, in the first row, the model cannot achieve a hollow structure by direct optimization because their topologies are inherently different. However, as shown in the second row, even though the topologies are different, the optimization method can still provide approximate results. For instance, the sphere-based initialization can shape the handle on the right side, even though the handle is incomplete. The sphere-based initialization can sometimes produce even more accurate results than our proposed method, as seen in the third row. These experiments demonstrate that our method is robust to initialization. Aiming for a better ability to generalize, we chose to use our fast initialization.

\section{User Study}

For user study, we render 360-degree videos of subject-driven 3D models and show each volunteer with five samples of rendered video from a random method. They can rate in four aspects: 3D consistency, subject fidelity, prompt fidelity, and overall quality on a scale of 1-10, with higher
scores indicating better performance. We collect results from 30 volunteers shown in Table~\ref{tab:user_study}. We find our method is significantly preferred by users over these aspects. 

\begin{table}[h]
    \centering
    \small
    \caption{Quantitative comparison results on the multi-view consistency, subject fidelity (related to geometric and texture details), prompt fidelity (related to the alignment of input single image), and overall quality score in a user study, rated on a range of 1-10, with higher scores indicating better performance.}
    \label{tab:user_study}
    \vspace{-1.5mm}
    \begin{tabular}{lcccc}
        \toprule
        Method  & Multi-view Consistency & Subject Fidelity & Prompt Fidelity & Overall Quality \\
        \midrule
        One-2-3-45~\cite{liu2024one-2-3-45} & 5.46 & 4.78 & 6.93 & 5.79 \\
        OpenLRM~\cite{LRM} & 6.72 & 7.16 & 6.92 & 7.15 \\
        SyncDreamer~\cite{liu2023syncdreamer} & 5.71 & 7.52 & 4.06 & 5.92 \\
        Wonder3D~\cite{long2023wonder3d} & 8.67 & 7.80 & 7.39 & 8.14 \\
        InstantMesh~\cite{xu2024instantmesh} & 8.31 & 7.68 & 7.91 & 8.43 \\
        GRM~\cite{GRM} & 6.93 & 7.42 & 6.02 & 7.38 \\
        CRM~\cite{CRM} & 7.95 & 8.53 & 8.03 & 8.25 \\
        \midrule
        \textbf{Ours} & \textbf{9.26} & \textbf{8.74} & \textbf{8.52} & \textbf{9.02} \\
        \bottomrule
    \end{tabular}
\end{table}

\section{Social Impact}

\textbf{Positive Impacts:}
The Unique3D framework can democratize 3D content creation, making it easier for artists and designers to produce 3D models from single images, which can lead to increased innovation and a surge in creative applications across various industries including gaming, film, and education.

\textbf{Negative Impacts:}
On the flip side, the ease of generating high-quality 3D models raises concerns about potential misuse, such as creating deepfakes, and could lead to job displacement for traditional 3D modelers. Additionally, there may be challenges related to intellectual property and privacy if the technology is used irresponsibly.

\section{Licenses for Used Assets}

Stable Diffusion~\cite{LDM} is under CreativeML Open RAIL M License.

Objaverse~\cite{Objaverse} is under ODC-By v1.0 license.